\documentclass{article}
\usepackage{ijcai25}

\usepackage{times}
\usepackage{soul}
\usepackage{url}
\usepackage[hidelinks]{hyperref}
\usepackage[utf8]{inputenc}
\usepackage[small]{caption}
\usepackage{graphicx}
\usepackage{amsmath}
\usepackage{amsthm}
\usepackage{booktabs}
\usepackage{algorithm}
\usepackage{algorithmic}
\usepackage[switch]{lineno}
\usepackage{array}

\usepackage{amsfonts}   
\usepackage{mathrsfs}

\title{Integration of Old and New Knowledge for Generalized Intent Discovery: \\A Consistency-driven Prototype-Prompting Framework}

\author{
Xiao Wei$^{1, 2, 3}$ \and
Xiaobao Wang$^{1, 2, *}$  \and
Ning Zhuang$^1$\and
Chenyang Wang$^4$\and\\
Longbiao Wang$^{1, 5, *}$ \And
Jianwu Dang$^3$\\
\affiliations
$^1$Tianjin Key Laboratory of Cognitive Computing and Application, \\College of Intelligence and Computing, Tianjin University, Tianjin, China\\
$^2$Guangdong Laboratory of Artificial Intelligence and Digital Economy (SZ), Shenzhen, China\\
$^3$Shenzhen Institute of Advanced Technology, Chinese Academy of Sciences, Shenzhen, China\\
$^4$College of Computer Science and Software Engineering, Shenzhen University, Shenzhen, China\\
$^5$Huiyan Technology (Tianjin) Co., Ltd, Tianjin, China\\
\emails
\{weixiao, wangxiaobao, zncola, longbiao\_wang\}@tju.edu.cn,\\
chenyangwang@ieee.org, jw.dang@siat.ac.cn
}

\begin{document}

\maketitle

\renewcommand{\thefootnote}{\fnsymbol{footnote}} 
\footnotetext[1]{Corresponding authors.} 

\begin{abstract}
Intent detection aims to identify user intents from natural language inputs, where supervised methods rely heavily on labeled in-domain (IND) data and struggle with out-of-domain (OOD) intents, limiting their practical applicability. Generalized Intent Discovery (GID) addresses this by leveraging unlabeled OOD data to discover new intents without additional annotation. However, existing methods focus solely on clustering unsupervised data while neglecting domain adaptation. Therefore, we propose a consistency-driven prototype-prompting framework for GID from the perspective of integrating old and new knowledge, which includes a prototype-prompting framework for transferring old knowledge from external sources, and a hierarchical consistency constraint for learning new knowledge from target domains. We conducted extensive experiments and the results show that our method significantly outperforms all baseline methods, achieving state-of-the-art results, which strongly demonstrates the effectiveness and generalization of our methods. Our source code is publicly available at \url{https://github.com/smileix/cpp}.
\end{abstract}

\section{Introduction}
Intent detection is a core task in both Natural Language Understanding (NLU) and Task-Oriented Dialogue (ToD) systems . Its primary goal is to identify the intent or objective of a user from their natural language input. Intent detection is a critical step for dialogue systems to understand user needs and take appropriate actions. In recent years, with the rise of deep learning, data-driven fully supervised methods have achieved significant success. However, these methods largely rely on a substantial amount of in-domain annotated data and can only handle a limited set of in-domain intents \cite{yang2024generalized}, which poses numerous challenges in practical applications. Generally, users will always inquire about some out of scope intents after the deployment of a dialogue system, which traditional dialogue systems may fail to correctly identify and respond to \cite{siddique2021linguistically}. 

\begin{figure}[tbp]
	\centering
    \includegraphics[width=1\linewidth]{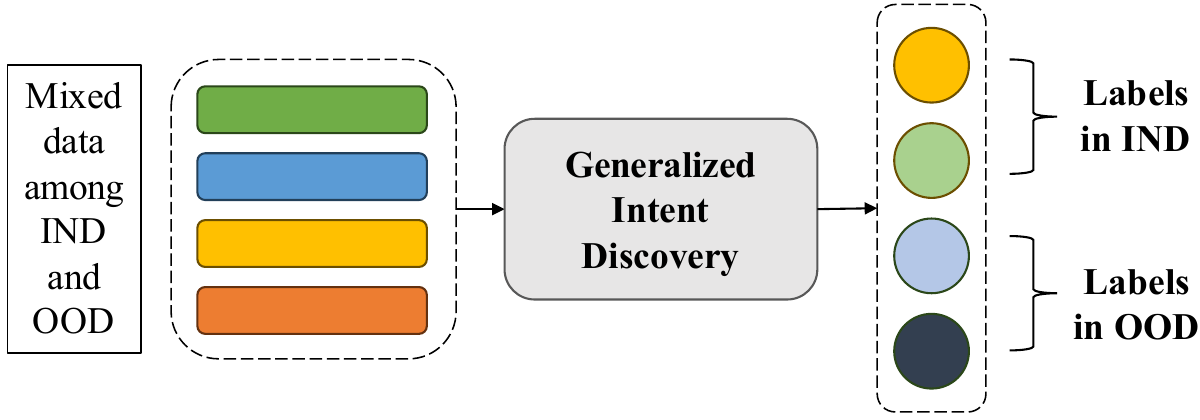}
	\caption{Illustration of the GID task. GID uses labeled IND data and unlabeled OOD data for training, and then jointly classifies IND and OOD data for evaluating.}
 		\label{fig:GID}
\end{figure}

Therefore, to enhance the generalization, scalability, and adaptability of dialogue systems, it is necessary to study Generalized Intent Discovery (GID) \cite{mou2022generalized,mou2023decoupling}. As the Fig.\ref{fig:GID} shows, GID uses labeled in-domain (IND) data and unlabeled out-of-domain (OOD) data for training, which aims to extend the traditional closed-domain intent set by leveraging such OOD data generated by dialogue systems. This reduces reliance on supervised data and gradually expands the system's intent detection scope, enabling it to recognize both known IND intents and unknown OOD intents, thereby improving the adaptability and intelligence of dialogue systems.

The GID task is an emerging research direction in task-oriented dialogue systems, whose core challenge lying in accurately classifying OOD intents in the absence of corresponding supervised data. Current mainstream methods are based on the EM algorithm \cite{dempster1977maximum}, which uses clustering methods to assign pseudo labels to unsupervised OOD data (E-Step) and then trains a joint classifier with supervised IND data (M-Step), repeating this process until the model converges. However, these methods focus on learning new knowledge in target domains while neglecting the transfer of knowledge from external sources \cite{wang2022generalizing,xiaobao1,xiaobao2}. This is manifested in two aspects. First, previous methods did not fully utilize prior knowledge when generating pseudo labels, resulting in low quality pseudo labels, introducing noise into the joint training phase  and limiting the performance of the model. Second, the fine-tuning paradigm is used during model training, which cannot fully eliminate the gap between pretraining tasks and downstream tasks due to insufficient supervised data in low resource scenarios, thus inadequately transferring knowledge from the pre-trained model.

To address these challenges, we propose a \textbf{C}onsistency-driven \textbf{P}rototype-\textbf{P}rompting framework (\textbf{CPP}) for GID,  from the perspective of integrating old and new knowledge. Our work makes three main contributions: 
\begin{itemize}
\item A prototype-prompt framework for transferring old knowledge from external sources, including prototype-optimized metric learning and prior-enhanced prompt learning. We exploit large language models (LLMs) to introduce different external knowledge in prototypes and verbalizers, and explore different update methods to improve the utilization of external knowledge.
\item Hierarchical consistency constraints for acquiring new knowledge from target domains, including consistency regularization and symmetric cross-prediction loss. We follow and extend the consistency principle to construct different constraints, which can also synergistically conduct metric learning and prompt learning.
\item Extensive experiments on up to 9 settings with different domain setups on two datasets, whose results show that our method significantly outperforms all baselines and achieves state-of-the-art results, proving the effectiveness and generalization of our method in generalized intent discovery.
\end{itemize}

\section{Related Work}
\paragraph{OOD Intent Detection \& Discovery.}
There are two tasks related to GID that are also designed to address OOD intent. OOD intent detection aims to identify whether a user query belongs to the predefined IND intent set and to reject OOD intents \cite{lang2023survey,zheng2020out}. However, the task cannot further utilize the potential information in these OOD queries. OOD intent discovery aims to cluster unlabeled OOD data and discover new intent categories \cite{vedula2019towards}, iteratively optimizing intent representations and cluster assignments \cite{lee2013pseudo}. However, it cannot effectively integrate with existing IND intent classifiers. The former is a fully supervised task, whose main methods includes distance-based detection \cite{xu2020distance}, generative model-based detection \cite{xu2021generative}, and contrastive learning-based detection \cite{zhou2022contrastive}. The latter is an unsupervised task, whose main methods includes K-means clustering \cite{lloyd1982least}, deep alignment-based clustering \cite{mou2022disentangled}, and contrastive learning-based clustering \cite{zhou2022knn}.
\paragraph{OOD Intent Detection \& Discovery.} GID is a semi-supervised task, where the current challenge lies in the high coupling between OOD pseudo label generation and representation learning \cite{sohn2020fixmatch,mou2023decoupling}. Specifically, the quality of  pseudo labels affects the performance of subsequent joint representation learning, while the effectiveness of representation learning influences the quality of pseudo label generation in turn.

Existing GID methods can be divided into two categories: pipeline or end-to-end. Pipeline methods are unidirectional, leading to error propagation and thus limiting model performance. Current mainstream methods are based on end-to-end approaches, such as E2E \cite{mou2022generalized} and DPL \cite{mou2023decoupling}. E2E introduces a framework that combines pseudo label generation and representation learning for joint optimization. Additionally, it employs swap prediction for OOD data to prevent the model from collapsing into degenerate solutions during clustering. Based on the end-to-end framework, DPL further introduces contrastive learning to enhance feature representation learning for OOD data in the target domain.
In summary, existing methods focus on optimizing OOD data clustering algorithms while neglecting domain adaption \cite{farahani2021brief} from external sources.

\section{Consistency-driven Prototype-Prompting Framework}

\subsection{Problem Formulation}

\paragraph{Input Data.} Labeled IND data: \( \textbf{D}_{\text{IND}} = \{(x_i^{\text{IND}}, y_i^{\text{IND}})\}_{i=1}^n \), where \( y_i^{\text{IND}} \in \mathcal{Y}_{\text{IND}} \), $\mathcal{Y}_{\text{IND}}$ is known and $|\mathcal{Y}_{\text{IND}}| = N$; unlabeled OOD data : \( \textbf{D}_{\text{OOD}} = \{(x_i^{\text{OOD}})\}_{i=1}^m \), where \( y_i^{\text{OOD}}\) is unknown, while \(\mathcal{Y}_{\text{OOD}}\) is known and $|\mathcal{Y}_{\text{OOD}}| = M$. \(\mathcal{Y}_{\text{IND}} \cap\mathcal{Y}_{\text{OOD}} = \emptyset\).

\paragraph{Objective.}
Use $\textbf{D}_{\text{IND}}$ and $\textbf{D}_{\text{OOD}}$ to train a joint classifier to classify input queries into the total label set \(\mathcal{Y} = \mathcal{Y}_{\text{IND}} \cup \mathcal{Y}_{\text{OOD}}\).

The model training in this domain is generally divided into two stages: the pretrain stage, where the model performs fully supervised learning on the labeled IND data, and the discover stage, where the model undergoes joint training on the labeled IND data and unlabeled OOD data, and then both IND and OOD data are used for evaluation.

\subsection{Overview}

\begin{figure}[tbp]
	\centering
    \includegraphics[width=1\linewidth]{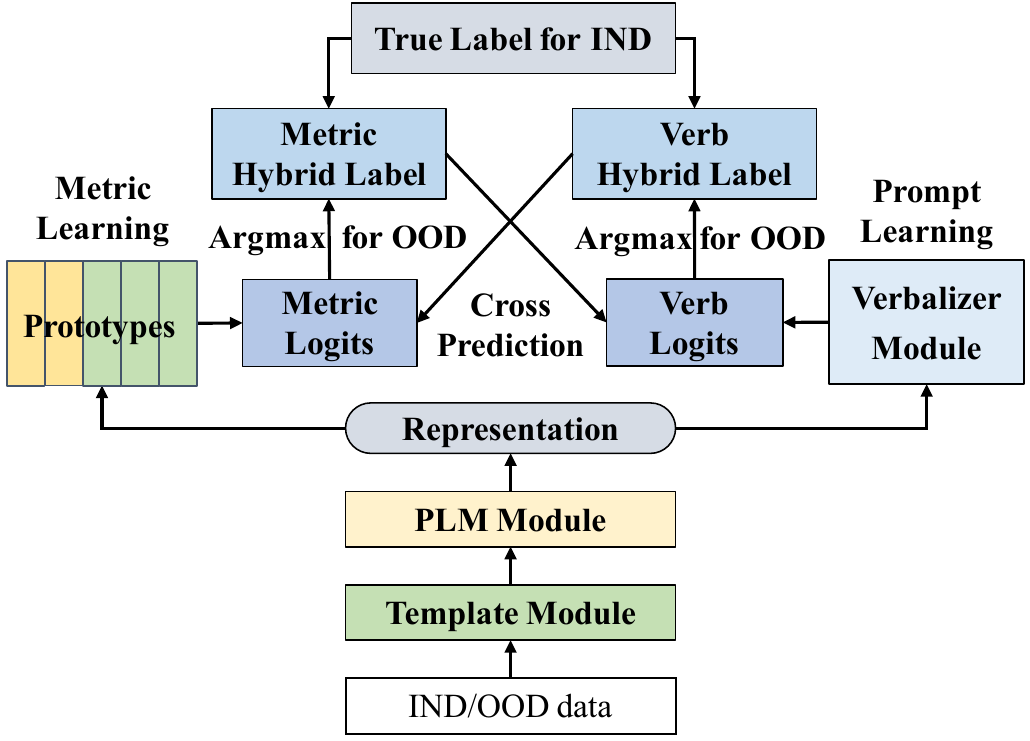}
	\caption{Overall architecture of our proposed method.}
 		\label{fig:model}
\end{figure}

The overall framework of our method is illustrated in the Figure \ref{fig:model}. For the discover stage, firstly, we design a prototype-prompt framework for transferring old knowledge from external sources, setting up two classifiers: a prototype-optimized similarity comparison module based on metric learning, and a prior-enhanced verbalizer layer based on prompt learning. Secondly, we construct hierarchical consistency constraints for acquiring new knowledge from target domains, and first set up consistency regularization. And then, we extend the idea of consistency principle and construct symmetric swapping prediction loss.

\subsection{Prototype-Prompting Framework for Old Knowledge Transfer}

\subsubsection{Prototype-Optimized Metric Learning}
Prototype-based metric learning \cite{kaya2019deep,wei22f_interspeech} aims to classify or cluster data by learning a prototype representation for each category and measuring the distance between samples and prototypes, which is particularly effective in low-resource scenarios.

We propose external knowledge-enhanced prototype metric learning based on LLMs, whose core idea is to leverage the semantic understanding capabilities of LLMs to generate high-quality initial prototypes for each category. 
We experiment with category names, descriptions, representative samples, and keywords. 

For prototype computation, specifically, we use the category name as the basic prototype text, denoted as \(\mathscr{C} = \{c_1, c_2, \dots, c_C\}\). We predefine a template and use a LLM (here, DeepSeek-V3) to generate corresponding meta-information denoted as $m_i$ like explanatory descriptions, representative samples, or keywords \cite{yuhang,wei2024prompt}:
\begin{equation}
m_i = \text{LLM}(t(c_i)), \label{eq:llm}
\end{equation}
where $t(\cdot)$ denote template function, including LLM role definition, dataset background and task description, input and output format description. And then we extract the corresponding embedding using our encoder denoted as $E(\cdot)$ to obtain the prototype for that category. Finally, we concatenate the prototype vectors of all categories to form the overall prototype vector:
\begin{align}
\mathbf{P} &= [E({m}_1); E({m}_2); \dots; E({m}_C)]. \label{eq:prototype}
\end{align}

For distance measurement, we compare the cosine similarity between the sample representation and the prototype representation to obtain its logits. For each training step, we repeat Formula \eqref{eq:prototype}  to update the prototype representation simply and effectively.

\subsubsection{Prior-Enhanced Prompting Framework}
Prompt learning aims to reduce the gap between pre-training tasks and downstream tasks, especially in low data scenarios, typically by restructuring downstream tasks into self-supervised tasks of pretrained models \cite{brown2020language,liu2023pre,ding2022openprompt}. Based on prompt learning, we predefine a template to reformulate the task into the form of Masked Language Model (MLM) \cite{devlin2019bert} head,  guide pretrained models to generate the expected output and design a verbalizer for downstream adaption \cite{zhuangning,junlei}, thus fully transferring the knowledge of the pretrained models. The framework consists of two main components: template and verbalizer, where we focus on the latter.

For an input text \( x \), we embed it into a template \( T(x) \):
\[
T(x) = \text{``}x\text{. In this sentence, the intent is about [MASK].''},
\]
where `[MASK]' is the position where the model predicts the label word. We use handcrafted templates without any trainable parameters.

The verbalizer acts as a function that maps the model's predicted label words to actual labels. It plays a crucial role in transferring prior knowledge in two ways: label words and MLM head reuse.

The construction of label words is similar to the initial text of prototypes, therefore we reuse the label-related meta-information generated from the LLM in Formula \eqref{eq:llm} as label words,  Let \( W_i \) denote the set of label words associated with the category label \( y_i \), and \( W_i = m_i\). Assuming each category \( y_i \) has \( k \) label words, \( W_i \) can be also expressed as $W_i = \{w_{i1}, w_{i2}, \dots, w_{ik}\}$.

We construct a soft verbalizer to reuse the MLM head more efficiently. Concretely, we introduce a downstream linear layer \( \mathbf{W}_{\text{verbalizer}} \in \mathbb{R}^{d \times C} \), where \( d \) is the embedding dimension and \( C \) is the number of categories. The weights of this layer are initialized by a subset of the MLM head's weights. The steps are as follows:

For each category label \( y_i \), we extract the corresponding weights from the MLM head's weight matrix \( \mathbf{W}_{\text{MLM}} \) based on its label words \( W_i \):
\begin{equation}
\mathbf{W}_{\text{verbalizer}}[:, i] = \frac{1}{k} \sum_{j=1}^k \mathbf{W}_{\text{MLM}}[:, w_{ij}]
\end{equation}
where \( \mathbf{W}_{\text{verbalizer}}[:, i] \) is the weight column for label \( y_i \) in the downstream linear layer, and \( w_{ij} \) represents the \( j \)-th label word for \( y_i \). And then we compute the logits for each label:
   
\begin{equation}
   S_i = \mathbf{W}_{\text{verbalizer}}[:, i]^\top h_{\text{[MASK]}},
\end{equation}
where \( h_{\text{[MASK]}} \) is the hidden representation of the `[MASK]` token. Finally,  the logits is normalized to obtain the probability distribution:
\begin{equation}
   p(y_i | x) = \frac{\exp(S_i)}{\sum_{j=1}^C \exp(S_j)}
\end{equation}

\subsection{Hierarchical Consistency Constraints for New Knowledge Acquisition}
In unsupervised learning scenarios, consistency regularization \cite{sohn2020fixmatch}, is a crucial technique that leverages the inherent properties of data to train models without relying on labeled data. The core idea is that for reasonable perturbations of the input data, the model's predictions should remain consistent. This allows the model to learn effective feature representations from the intrinsic structure of the data.

\subsubsection{Consistency Regularization Loss}
We propose a consistency regularization (CR) \cite{fan2023revisiting} loss based on the principle of consistency, including data consistency and prediction consistency.

As for the data consistency, we generate two different representation views of the same input with the dropout mechanism \cite{little1995modeling}, denoted as $\textbf{H}_1$ and $\textbf{H}_2$. And then we use symmetric KL divergence \cite{kullback1951information} to align the distributions of these two views:
\begin{equation}
     \mathcal{L}_{\text{dc}} = \frac{1}{2} \left( D_{\text{KL}}(\textbf{H}_1 \| \textbf{H}_2) + D_{\text{KL}}(\textbf{H}_2 \| \textbf{H}_1) \right),
\end{equation}
where \( \textbf{H}_1 \) and \( \textbf{H}_2 \) are obtained by applying mean pooling to the final hidden states of the encoder.

As for the prediction consistency, the two classifiers, denoted as $C_1$ and $C_2$, process the same input and obtain two probability distributions for each view. And then we again use symmetric KL divergence to align these distributions:
\begin{equation}
\textbf{P}_1 = C_1(\textbf{H}), \quad \textbf{P}_2 = C_2(\textbf{H}),
\end{equation}
\begin{equation}
     \mathcal{L}_{\text{pc}} = \frac{1}{2} \left( D_{\text{KL}}(\textbf{P}_1 \| \textbf{P}_2) + D_{\text{KL}}(\textbf{P}_2 \| \textbf{P}_1) \right),
\end{equation}
where \( \textbf{P}_1 \) and \( \textbf{P}_2 \) are the probability distributions from the two classifiers. The final consistency regularization loss is the sum of these two losses:
\begin{equation}
     \mathcal{L}_{\text{CR}} =  \mathcal{L}_{\text{dc}} +  \mathcal{L}_{\text{pc}}.
\end{equation}

\subsubsection{Symmetric Cross-Prediction Loss}
Since GID is a semi-supervised task, where unsupervised data lacks labels during training. To address this, we propose a symmetric cross-prediction loss, based on the two classifiers we constructed. Specifically, for each view:\\
1). Take argmax of the logits to generate pseudo labels for OOD data:
\begin{equation}
    \hat{Y}_1 = \arg\max(\textbf{P}_1), \quad \hat{Y}_2 = \arg\max(\textbf{P}_2),
\end{equation}
2). Combine these pseudo labels of OOD data with the true labels of IND data to form hybrid labels.
\\\begin{equation}
    \text{Hybrid}_1 = \begin{cases}
Y & \text{(IND)} \\
\hat{Y}_2 & \text{(OOD)}
\end{cases},
\text{Hybrid}_2 = \begin{cases}
Y & \text{(IND)} \\
\hat{Y}_1 & \text{(OOD)}
\end{cases},
\end{equation}
where the \( \text{Hybrid}_1 \) and \( \text{Hybrid}_2 \) are the hybrid labels.\\
3). Compute the cross-entropy (CE) loss \cite{shannon1948mathematical,goodfellow2016deep} with hybrid labels and the other classifier's logits, and then symmetrically calculate and accumulate it:
\begin{equation}
   \mathcal{L}_{\text{CP}} = \frac{1}{2} \left( \text{CE}(\textbf{P}_1, \text{Hybrid}_2) + \text{CE}(\textbf{P}_2, \text{Hybrid}_1) \right) .
\end{equation}

\subsubsection{Multi-View Contrastive Learning}
For unsupervised data, we also incorporate a classic contrastive learning approach \cite{chuang2020debiased}. Specifically, based on the two views mentioned above, we apply multi-view contrastive learning with the NT-Xent loss \cite{chen2020simple}. The goal is to maximize the similarity of positive pairs (views of the same data) and minimize the similarity of negative pairs (views of different data):
\begin{equation}
\mathcal{L}_{\text{CL}} = -\sum_{i=1}^{2B} \log \frac{\exp(\text{sim}(z_i, \hat{z_i})/\tau)}{\sum_{k=1}^{2B} \mathbb{I}_{k \neq i} \exp(\text{sim}(z_i, z_k)/\tau)}
\end{equation}
where \( z_i \) and \( \hat{z_i} \) are the feature representations of the two views of the same data (positive pair), which are obtained by projecting the averaged hidden states of last layer, \( \text{sim}(\cdot) \) represents a similarity function (here cosine similarity), $B$ is the batch size and \( \tau \) is a temperature parameter.

Our final loss is obtained by adding up the three losses:
\begin{equation}
     \mathcal{L} =  \mathcal{L}_{\text{CR}} +  \mathcal{L}_{\text{CP}}  +  \mathcal{L}_{\text{CL}}.
\end{equation}

\begin{table*}[ht!]
\centering
\small
\scalebox{0.93}{
        \begin{tabular}{l!{\vrule width 0.75pt}c|cc|cc!{\vrule width 0.75pt}c|cc|cc!{\vrule width 0.75pt}c|cc|cc}
        \toprule
        Setup & \multicolumn{5}{c!{\vrule width 0.75pt}}{GID-SD-60\%} & \multicolumn{5}{c!{\vrule width 0.75pt}}{GID-CD-60\%} & \multicolumn{5}{c}{GID-MD-60\%}\\
        \midrule
        Metric →& IND& \multicolumn{2}{c|}{OOD}& \multicolumn{2}{c!{\vrule width 0.75pt}}{ALL}& IND& \multicolumn{2}{c|}{OOD }& \multicolumn{2}{c!{\vrule width 0.75pt}}{ALL}& IND& \multicolumn{2}{c|}{OOD}& \multicolumn{2}{c}{ALL}\\
        Model \  ↓& ACC& ACC&    F1&  ACC&  F1&  ACC&  ACC&  F1&  ACC&   F1 &  ACC&  ACC&  F1&  ACC& F1 \\
        \midrule
        Kmeans& 90.40& 51.58& 51.96& 67.08& 66.70& 96.44 & 54.67 & 53.69 & 71.38 & 70.57 
& 95.00 & 65.11 & 63.68 & 77.02 &76.09 
\\
        DA& 90.97 & 59.55 & 59.51 & 72.05 & 71.42 
& 97.33 & 76.15 & 74.80 & 84.62 & 83.60 
& 97.67 & 83.38 & 82.78 & 89.10 &88.52 
\\
        DA-Mix& 80.70 & 52.66 & 54.66 & 63.95 & 61.92 
& 93.89 & 75.63 & 74.29 & 82.93 & 81.37 
& 92.59 & 78.34 & 79.88 & 84.05 &82.74 
\\
        E2E& 91.77& \underline{62.23}& \underline{62.62}& \underline{73.93}& \underline{73.77}& \textbf{98.67}& \underline{80.81}& \underline{80.60}& \underline{87.96}& \underline{87.68}& \underline{98.11}& \underline{87.19}& \underline{87.32}& \underline{91.51}&\underline{91.24}\\
        DPL& \underline{92.66}& 59.84& 60.29& 72.91& 72.54& 98.00& 79.78& 79.56& 87.07& 86.78& 97.93& 86.79& 87.01& 91.26&91.18\\
        \midrule
        \textbf{CPP}& \textbf{93.63}& \textbf{75.05}& \textbf{74.92}& \textbf{82.53}& \textbf{81.84}& \underline{98.11}& \textbf{87.41}& \textbf{86.72}& \textbf{91.69}& \textbf{91.17}& \textbf{98.44}& \textbf{90.96}& \textbf{91.02}& \textbf{93.96}&\textbf{93.73}\\
        \bottomrule
        \end{tabular}}       
\caption{Performance on the SD (Single Domain), CD (Cross Domain) and MD (Multi Domain) setups with 60\% OOD ratio. Here DA stands for DeepAligned, and DA-Mix stands for DeepAligned-Mix. For IND data, accuracy is reported, while for OOD data and overall data, accuracy and weighted-F1 score are reported. Note that the best value in each column is represented in bold, and the suboptimal value is represented by an underline. The same below.
}
\label{tbl:gid_60}
\end{table*}

\begin{table*}[ht]
\centering
\small
\scalebox{0.93}{
        \begin{tabular}{l!{\vrule width 0.75pt}c|cc|cc!{\vrule width 0.75pt}c|cc|cc!{\vrule width 0.75pt}c|cc|cc}
        \toprule
        Setup & \multicolumn{5}{c!{\vrule width 0.75pt}}{GID-SD-80\%} & \multicolumn{5}{c!{\vrule width 0.75pt}}{GID-CD-80\%} & \multicolumn{5}{c}{GID-MD-80\%}\\
        \midrule
        Metric →& IND& \multicolumn{2}{c|}{OOD}& \multicolumn{2}{c!{\vrule width 0.75pt}}{ALL}& IND& \multicolumn{2}{c|}{OOD }& \multicolumn{2}{c!{\vrule width 0.75pt}}{ALL}& IND& \multicolumn{2}{c|}{OOD}& \multicolumn{2}{c}{ALL}\\
        Model \  ↓& ACC& ACC&    F1&  ACC&  F1&  ACC&  ACC&  F1&  ACC&   F1 &  ACC&  ACC&  F1&  ACC& F1 \\
        \midrule
        E2E& 91.17 & \underline{57.98} & \underline{57.69} & \underline{64.42} & \underline{63.86} 
& 96.00 & \underline{84.39} & \underline{84.32} & \underline{86.71} & \underline{86.53} 
& 97.56 & \underline{74.78} & \underline{74.88} & \underline{79.33} &\underline{79.30} 
\\
        DPL& \textbf{91.94} & 50.94 & 50.78 & 58.90 & 58.44 
& \textbf{98.67} & 82.57 & 82.54 & 85.79 & 85.61 
& \textbf{98.00} & 74.11 & 73.66 & 78.71 &78.33 
\\
        \midrule
        \textbf{CPP}& \underline{91.17} & \textbf{66.85} & \textbf{64.42} & \textbf{71.59} & \textbf{69.23} 
& \underline{96.67} & \textbf{86.61} & \textbf{86.58} & \textbf{88.62} & \textbf{88.45} 
& \underline{97.56} & \textbf{79.61} & \textbf{79.18} & \textbf{83.20} &\textbf{82.89} 
\\
        \bottomrule
        \end{tabular}}        
\caption{Performance on the SD (Single Domain), CD (Cross Domain) and MD (Multi Domain) setups with 80\% OOD ratio. }
\label{tbl:gid_80}
\end{table*}

\begin{table*}[ht]
\centering
\small
\scalebox{0.93}{
        \begin{tabular}{l!{\vrule width 0.75pt}c|cc|cc!{\vrule width 0.75pt}c|cc|cc!{\vrule width 0.75pt}c|cc|cc}
        \toprule
        Setup & \multicolumn{5}{c!{\vrule width 0.75pt}}{GID-SD-90\%} & \multicolumn{5}{c!{\vrule width 0.75pt}}{GID-CD-90\%} & \multicolumn{5}{c}{GID-MD-90\%}\\
        \midrule
        Metric →& IND& \multicolumn{2}{c|}{OOD}& \multicolumn{2}{c!{\vrule width 0.75pt}}{ALL}& IND& \multicolumn{2}{c|}{OOD }& \multicolumn{2}{c!{\vrule width 0.75pt}}{ALL}& IND& \multicolumn{2}{c|}{OOD}& \multicolumn{2}{c}{ALL}\\
        Model \  ↓& ACC& ACC&    F1&  ACC&  F1&  ACC&  ACC&  F1&  ACC&   F1 &  ACC&  ACC&  F1&  ACC& F1 \\
        \midrule
        E2E& 89.69 & \underline{52.28} & \underline{51.87} & \underline{56.17} & \underline{55.63} 
& \underline{96.89} & \underline{75.95} & \underline{75.57} & \underline{78.00} & \underline{77.53} 
& 96.44 & 53.88 & 53.50 & 58.13 &57.84 
\\
        DPL& \textbf{90.42} & 40.04 & 39.20 & 45.14 & 44.15 
& 96.89 & 75.85 & 75.63 & 77.93 & 77.66 
& \underline{96.44} & \underline{67.51} & \underline{67.14} & \underline{70.49} &\underline{70.17} 
\\
        \midrule
        \textbf{CPP}& \underline{87.19} & \textbf{63.62} & \textbf{61.21} & \textbf{66.07} & \textbf{63.67} 
& \textbf{98.22} & \textbf{83.51} & \textbf{82.73} & \textbf{84.98} & \textbf{84.20} 
& \textbf{96.44} & \textbf{82.17} & \textbf{81.63} & \textbf{83.60} &\textbf{83.13} 
\\
        \bottomrule
        \end{tabular}}     
\caption{Performance on the SD (Single Domain), CD (Cross Domain) and MD (Multi Domain) setups with 90\% OOD ratio. }
\label{tbl:gid_90}
\end{table*}

\section{Experimental Results}
\subsection{Datasets and Settings}
We conducted experiments on two classic intent detection datasets: Banking \cite{casanueva2020efficient} and CLINC \cite{larson2019evaluation}. Banking is specifically designed for banking scenarios, containing 77 different banking-related intents with approximately 13,000 samples. CLINC covers 10 domains and includes 150 different intent categories, with approximately 22,500 samples.

We adopted the experimental setups from previous work \cite{mou2022generalized,mou2023decoupling} and further extended data settings . 
We followed the setups of GID-SD, GID-CD, and GID-MD, which means single-domain, cross-domain and multi-domain respectively. Specifically, GID-SD refers to extracting a certain proportion of data as OOD unsupervised data on Banking, with intent categories as the smallest partition unit, GID-MD refers to performing the same operation on CLINC, while GID-CD refers to obtain such data on CLINC with domain as the smallest partition unit. The remaining data is used as IND supervised data for all the three setups. The model is required to use both IND  data with label and OOD data without labels for joint training, and then is expected to predict both intents of IND and OOD. For each setup, previous work used OOD ratios of 20\%, 40\%, and 60\%. We follow the 60\% OOD ratio and expand it to 80\% and 90\%, make it more challenging and discriminative.

\subsection{Baselines}
We compared our method with the following  baselines:
\paragraph{Kmeans} \cite{lloyd1982least}: A pipeline approach that first clusters OOD data with k-means to generate pseudo labels and then jointly trains with IND data.
\paragraph{DeepAligned} \cite{zhang2021discovering}: A pipeline approach that improves upon Kmeans by using the Hungarian algorithm to align cluster centers after clustering, addressing label inconsistency.
\paragraph{DeepAligned-Mix} \cite{mou2022generalized}: A variant of DeepAligned that differs mainly in the inference phase, where predictions are made using the classification head of the new classifier.
\paragraph{E2E} \cite{mou2022generalized}: An end-to-end approach for pseudo label generation and intent classification, using Sinkhorn-Knopp (SK) \cite{sinkhorn1967concerning} algorithm to optimize pseudo label assignment and employing multi-view swap prediction \cite{chen2020simple}.
\paragraph{DPL} \cite{mou2023decoupling}: An end-to-end approach that builds on E2E, using prototype contrastive learning to learn and generate pseudo labels for OOD data. It achieved state-of-the-art results on the previous 40\% OOD ratio settings.

For fair comparison, all baselines use BERT (bert-base-uncased) \cite{devlin2019bert} as the encoders, and are evaluated in terms of accuracy and weighted F1-score follow previous settings.

\subsection{Implementation Details} 
All text is tokenized with wordpieces, the maximum sequence length is set to 128, and dynamic padding is used; the batch size is set to 64, the maximum training epoch is set to 25, and the early stop is set to 10; the bert-base-uncased pretrained model is loaded as the backbone, and the dropout is set to 0.1; the optimizer is set to AdamW \cite{loshchilov2017decoupled}, the weight decay is 0.01, the initial learning rate is set to 5e-5, and a linear warm-up strategy \cite{goyal2017accurate} with 500 warm-up steps is adopted; we utilized prompt-based fine-tuning and set up an additional projection layer with 256 dimensional for NT-Xent loss. To ensure comparability, we directly use the code released in previous studies, especially E2E and DPL. All of the experiments were conducted on 8 NVIDIA RTX 4090D GPUs.

\subsection{Main Results}   
We conducted extensive experiments for the three setups (GID-SD, GID-CD, and GID-MD) with three data settings (60\%, 80\%, and 90\% OOD ratios) . We compared the performance with baselines, and the results are shown in Tables \ref{tbl:gid_60}, \ref{tbl:gid_80}, and \ref{tbl:gid_90}. Following convention, we use 5 evaluation metrics for each experiment, i.e., accuracy of IND data, accuracy and weighted-F1 score of OOD data, and overall accuracy and weighted-F1 score. Note that for the experimental results not provided by the baselines themselves, we reproduced them via their source codes.

The experimental results show that our method (CPP) significantly outperforms the baselines across all settings, achieving state-of-the-art performance. Overall, compared to the suboptimal model and in terms of total accuracy, our method leads by 8.60\%, 7.17\%, and 9.90\%, respectively for the 60\%, 80\%, and 90\% OOD ratios in the SD setup. In the CD setup, these numbers are 3.73\%, 3.87\%, and 13.11\%, and in the MD setup, they are 2.45\%, 1.91\%, and 6.98\%. These results strongly demonstrate the effectiveness and generalization of our method. Further observation reveals that our method achieves the best results on almost all metrics (40/45). The performance of different methods on IND data is roughly comparable, while our improvement on OOD data is more critical. Baseline methods tend to focus on clustering learning for unsupervised data while neglecting domain adaptation. Our improvement is attributed to the old-new knowledge fusion mechanism, which effectively transfers external knowledge and source domain knowledge to target domains while learning new knowledge from unsupervised data through hierarchical consistency constraints.

We conducted experiments on three setups: SD, CD, and MD. The performance of all methods gradually improves across these setups. This is because, in the SD setup, the model needs to recognize multiple fine-grained intents within a single domain, increasing the classification challenge. Compared to the MD setup, the CD setup makes it difficult for the model to transfer shared knowledge from source domains to the target domains due to the cross-domain setting between IND and OOD data. The improvement of our method over the suboptimal methods in these three settings generally follows this trend. For example, for the 60\% OOD ratio, our method improves over E2E by 8.60\%, 3.73\%, and 2.45\%, respectively. Additionally, we evaluated three OOD ratios: 60\%, 80\%, and 90\%. The performance of all methods gradually declines as the OOD ratio increases, which aligns with our intuition. The most significant improvement of our method over the suboptimal model occurs at the 90\% OOD ratio. The larger the OOD ratio, the more urgent the need for relevant knowledge in GID methods. For example, in the GID-CD-90\% setup, our method improves over DPL by 13.11\%, demonstrating that even in extremely resource-scarce scenarios, our method can effectively transfer old knowledge from external sources and learn new knowledge from OOD data.
 
\begin{table}[tbp!]
\centering
\small
\scalebox{0.95}{
        \begin{tabular}{l!{\vrule width 0.75pt}c|cc|cc}
        \toprule
        Setup & \multicolumn{5}{c}{GID-SD-60\%}  \\
        \midrule
        Metric →& IND& \multicolumn{2}{c|}{OOD}& \multicolumn{2}{c}{ALL}\\
        Model \  ↓& ACC& ACC&    F1&  ACC&  F1\\
        \midrule
        0-Shot DeepSeek& 77.42 & 69.13 & 73.93 & 72.47 & 72.12 
\\
        0-Shot GPT-4o& 72.81 & 72.90 & 73.26 & 72.89 & 72.10 
\\
        1-Shot DeepSeek& \underline{87.14}& 72.55 & \textbf{76.49} & \underline{78.51} & \underline{77.88} 
\\
        1-Shot GPT-4o& 83.95 & \underline{73.37} & \underline{76.79} & 77.63 & 77.56 
\\
        \midrule
        \textbf{CPP}& \textbf{93.63} & \textbf{75.05} & 74.92 & \textbf{82.53} & \textbf{81.84} 
\\
        \bottomrule
        \end{tabular}}       
\caption{Metric on GID-SD-60\% for comparison with LLMs.}
        
\label{tbl:llm}
\end{table}

\section{Quantitative Analysis}
\subsection{Comparison with LLMs}
In the field of Natural Language Processing (NLP), LLMs have demonstrated powerful performance across a variety of tasks. However, these models typically require substantial computational resources and data, which limits their applicability in resource-constrained scenarios. 
Our approach integrates both large and small language models (SLMs), leveraging the strengths of LLMs to assist SLMs in decision-making, thereby combining their complementary advantages to achieve low resource consumption and high performance.

For further analysis, we select the recently popular DeepSeek-V3 and the established GPT-4o as baselines, and conduct zero-shot and few-shot experiments. 
Note that 1-shot refers that for each IND category, one  random sample with the corresponding label is used , while for OOD categories, $|OOD * 1|$ samples are randomly chosen from the OOD training data in total, due to the absence of their labels.

As shown in Table \ref{tbl:llm}, our method significantly outperforms these LLMs in terms of overall metrics. The accuracy of our method is higher than that of DeepSeek 1-shot by 4.02\% and higher than GPT-4o 1-Shot by 4.90\%. In a closer observation, this improvement is primarily attributed to the strong performance on IND data, where our method surpasses DeepSeek 1-shot by 6.49\% and GPT-4o 1-Shot by 9.68\%. In contrast, the gap narrows significantly on OOD data, where our method maintains a slight lead in accuracy but falls slightly behind in F1 scores. This suggests that our method still have potential for handling of OOD data. Overall, our method demonstrates superior performance on specific tasks compared to LLMs, particularly in resource-constrained scenarios, making it a competitive choice. 
\begin{table}[tbp]
\centering
\small
\scalebox{1}{
        \begin{tabular}{l!{\vrule width 0.75pt}c|cc|cc}
        \toprule
        Setup & \multicolumn{5}{c}{GID-SD-60\%}  \\
        \midrule
        Metric →& IND& \multicolumn{2}{c|}{OOD}& \multicolumn{2}{c}{ALL}\\
        Setting \  ↓& ACC& ACC&    F1&  ACC&  F1\\
        \midrule
        Label name& 92.42 & 58.97 & 59.49 & 72.44 & 72.34 
\\
        Paraphrase& \textbf{93.71} & 57.99 & 55.31 & 72.37 & 70.05 
\\
         Keywords& 90.16 & 74.29 & 75.29 & 80.68 & 80.39 
\\
        1 example& 93.47 & 71.52 & 71.85 & 80.36 & 79.88 
\\
 2 examples& 93.23 & 73.59 & 73.79 & 81.49 &81.03 
\\
 3 examples& 93.55 & \underline{75.98} & \underline{76.25} & \underline{83.05} &\underline{82.56} 
\\
 4 examples& 93.55 & \textbf{78.64} & \textbf{79.61} & \textbf{84.64} &\textbf{84.56} 
\\
        \midrule
        5 examples& \underline{93.63} & 75.05 & 74.92 & 82.53 & 81.84 
\\
        \bottomrule
        \end{tabular}}
\caption{Metric on GID-SD-60\% with various Meta-information.}
        
\label{tbl:meta-information}
\end{table}

\subsection{Utilization of Meta-information}
In few-shot learning and zero-shot learning scenarios, the performance of SLMs is often limited by their constrained parameter scale and training data. To enhance the performance of small models in complex tasks, an effective approach is to leverage meta-information related to labels generated by LLMs to augment the decision-making capabilities of small models. This method not only utilizes the powerful semantic understanding abilities of large models but also compensates for the shortcomings of small models in data-scarce situations through knowledge transfer. We conduct this method by injecting label-related descriptions, such as paraphrases, keywords, and examples, into the prototype and verbalizer of our SLM. As shown in Table \ref{tbl:meta-information}, the experimental results demonstrate that this method significantly improves the performance of small models, highlighting its important application value in resource-constrained scenarios.

Specifically, the performance across all configurations on IND data remains relatively stable, indicating that the enhancement effect of meta-information on IND data is limited, with the primary improvement observed on OOD data. When only label names or paraphrases are used, the performance on OOD data is relatively low, suggesting that they provide limited semantic information. Keywords and examples are more effective. In the GID-SD-60\% setup, as examples increases, the performance on OOD data improves significantly, reaching its peak when using 4 examples. This suggests that a greater number of examples can provide richer contextual information. However, when examples increases to 5, the performance slightly declines. It is worth noting that we conducted this experiment across all other setups as well, and the conclusions vary slightly across different setups. In 6 out of 9 setups, using 5 examples yields the best performance.

 \begin{table}[tbp]
\centering
\small
\scalebox{1}{
        \begin{tabular}{l!{\vrule width 0.75pt}c|cc|cc}
        \toprule
        Setup & \multicolumn{5}{c}{GID-SD-60\%}  \\
        \midrule
        Metric →& IND& \multicolumn{2}{c|}{OOD}& \multicolumn{2}{c}{ALL}\\
        Setting \  ↓& ACC& ACC&    F1&  ACC&  F1\\
        \midrule
        Soft Template
& 92.90 & \underline{71.09} & \underline{71.28} & \underline{79.87} & \underline{78.98} 
\\
        PLM Freeze v1
& 66.85 & 34.35 & 34.48 & 47.44 & 42.11 
\\
         PLM Freeze v2
& \underline{93.55} & 60.16 & 57.17 & 73.60 & 71.00 
\\
        Manual Verb& 60.08 & 39.67 & 35.62 & 47.89 & 39.52 
\\
 PTR Verbalizer
& 59.76 & 40.16 & 36.88 & 48.05 &40.01 
\\
        \midrule
        \textbf{CPP}
& \textbf{93.63} & \textbf{75.05} & \textbf{74.92} & \textbf{82.53} & \textbf{81.84} 
\\
        \bottomrule
        \end{tabular}}
\caption{Metric on GID-SD-60\% in various parameter settings.}
        
\label{tbl:parameter}
\end{table}

\subsection{Ablation Study}
\subsubsection{Parameter Settings}
Prompt learning and parameter-efficient fine-tuning have emerged as highly regarded research directions in recent years. The effectiveness of prompt learning heavily depends on the parameter settings. Therefore, we first conducted parameter update experiments, as shown in Table \ref{tbl:parameter}. We systematically explore the impact of different parameter configurations on model performance, through ablation experiments on the parameter settings of our prompt framework, the encoder, and the verbalizer. PLM Freeze v1 refers to freezing all parameters in the pretrained language model (PLM), while PLM Freeze v2 refers to unfreezing its last layer on the basis of v1.

Our method, which combines hard templates, prompt-based fine-tuning, and soft verbalizers, achieves the best performance. Upon closer examination,  although completely freezing the PLM parameters leads to a significant performance drop, the model still achieves an accuracy of 47.44\%, demonstrating that our framework is highly effective in transferring knowledge from external sources into target domains. Both manually designed and rule-based PTR \cite{han2021ptr} verbalizers perform poorly, indicating their lack of flexibility. In contrast, the soft verbalizer dynamically generates label word mappings, better capturing task semantics and significantly improving the model's performance on both IND and OOD data.
  
\begin{table}[tbp]
\centering
\small
\scalebox{1}{
        \begin{tabular}{l!{\vrule width 0.75pt}c|cc|cc}
        \toprule
        Setup & \multicolumn{5}{c}{GID-SD-60\%}  \\
        \midrule
        Metric →& IND& \multicolumn{2}{c|}{OOD}& \multicolumn{2}{c}{ALL}\\
        Setting \  ↓& ACC& ACC&    F1&  ACC&  F1\\
        \midrule
        w/o $\mathcal{L}_{dc}$& 93.55 & 73.70 & 74.11 & 81.69 & 80.97 
\\
        w/o $\mathcal{L}_{pc}$ & 92.42 & \underline{74.51} & \underline{74.11} & \underline{81.72} & \underline{80.98} 
\\
        w/o $\mathcal{L}_{CP}$ & 93.06 & 70.54 & 70.52 & 79.61 & 78.95 
\\
        w/o $\mathcal{L}_{CL}$ & \textbf{94.03} & 69.29 & 67.89 & 79.25 &77.84 
\\
        \midrule
        \textbf{CPP} & \underline{93.63} & \textbf{75.05} & \textbf{74.92} & \textbf{82.53} & \textbf{81.84} 
\\
        \bottomrule
        \end{tabular}}
\caption{Metric on GID-SD-60\% with various loss functions.}
        
\label{tbl:loss}
\end{table}

\subsubsection{Loss Function Settings}
To better acquire new knowledge from target domains, we designs three types of loss functions: consistency regularization loss (including data consistency loss and prediction consistency loss), symmetric cross-prediction loss, and contrastive learning loss. Each loss function plays a distinct role in model training. Through ablation experiments, we can clearly evaluate the specific contribution of each loss function to model performance, thereby validating their rationality and necessity. By analyzing the impact of each loss function on model performance, we can further optimize the weights of the loss functions or design more efficient training strategies.

As shown in Table \ref{tbl:loss}, the ablation experiment results demonstrate that each of these loss functions contributes to model performance to varying degrees. Specifically, removing the contrastive learning loss results in the most significant performance drop on OOD data, indicating its irreplaceable role in enhancing the feature representation of unlabeled data. Following that, the symmetric cross-prediction loss also shows its importance in improving the model's classification capability on unlabeled data. Additionally, the data consistency loss plays a certain role in enhancing the quality of representation learning, while the prediction consistency constraint helps reduce the negative impact of pseudo-label noise on the model.

\section{Conclusion}
In this study, we propose our method from the perspective of integrating old and new knowledge, which includes a prototype-prompt framework for transferring old knowledge from external sources and hierarchical consistency constraints for acquiring new knowledge in target domain. For the former, we introduce LLMs to generate label-related meta-information and apply it into metric learning and prompt learning to reinforce SLMs decisions, thereby achieving complementary advantages of large and small PLMs. For the latter, we synergistically conduct prompt learning and metric learning to further improve performance. We conducted extensive experiments and analysis on various settings to validate the effectiveness and generalization of our method. In the future, we will attempt to load more powerful LLMs to explore parameter-efficient fine-tuning methods in this scenario, further improving its performance under extremely low resource conditions.

\section*{Acknowledgments}
This work was supported by the National Natural Science Foundation of China (No. 62302333 and No. U23B2053) and the Open Research Fund from Guangdong Laboratory of Artificial Intelligence and Digital Economy (SZ) (No.GML-KF-24-16).

\bibliographystyle{named}
\bibliography{ijcai25}

\begin{thebibliography}{}

\bibitem[\protect\citeauthoryear{Brown \bgroup \em et al.\egroup }{2020}]{brown2020language}
Tom Brown, Benjamin Mann, Nick Ryder, Melanie Subbiah, Jared~D Kaplan, Prafulla Dhariwal, Arvind Neelakantan, Pranav Shyam, Girish Sastry, Amanda Askell, et~al.
\newblock Language models are few-shot learners.
\newblock {\em Advances in neural information processing systems}, 33:1877--1901, 2020.

\bibitem[\protect\citeauthoryear{Casanueva \bgroup \em et al.\egroup }{2020}]{casanueva2020efficient}
I{\~n}igo Casanueva, Tadas Tem{\v{c}}inas, Daniela Gerz, Matthew Henderson, and Ivan Vuli{\'c}.
\newblock Efficient intent detection with dual sentence encoders.
\newblock {\em arXiv preprint arXiv:2003.04807}, 2020.

\bibitem[\protect\citeauthoryear{Chen \bgroup \em et al.\egroup }{2020}]{chen2020simple}
Ting Chen, Simon Kornblith, Mohammad Norouzi, and Geoffrey Hinton.
\newblock A simple framework for contrastive learning of visual representations.
\newblock {\em arXiv preprint arXiv:2002.05709}, 2020.

\bibitem[\protect\citeauthoryear{Chuang \bgroup \em et al.\egroup }{2020}]{chuang2020debiased}
Ching-Yao Chuang, Joshua Robinson, Yen-Chen Lin, Antonio Torralba, and Stefanie Jegelka.
\newblock Debiased contrastive learning.
\newblock {\em Advances in neural information processing systems}, 33:8765--8775, 2020.

\bibitem[\protect\citeauthoryear{Dempster \bgroup \em et al.\egroup }{1977}]{dempster1977maximum}
Arthur~P Dempster, Nan~M Laird, and Donald~B Rubin.
\newblock Maximum likelihood from incomplete data via the em algorithm.
\newblock {\em Journal of the Royal Statistical Society: Series B (Methodological)}, 39(1):1--22, 1977.

\bibitem[\protect\citeauthoryear{Devlin \bgroup \em et al.\egroup }{2019}]{devlin2019bert}
Jacob Devlin, Ming-Wei Chang, Kenton Lee, and Kristina Toutanova.
\newblock Bert: Pre-training of deep bidirectional transformers for language understanding.
\newblock In {\em Proceedings of the 2019 Conference of the North American Chapter of the Association for Computational Linguistics: Human Language Technologies, Volume 1 (Long and Short Papers)}, pages 4171--4186, 2019.

\bibitem[\protect\citeauthoryear{Ding \bgroup \em et al.\egroup }{2022}]{ding2022openprompt}
Ning Ding, Shengding Hu, Weilin Zhao, Yulin Chen, Zhiyuan Liu, Haitao Zheng, and Maosong Sun.
\newblock Openprompt: An open-source framework for prompt-learning.
\newblock In {\em Proceedings of the 60th Annual Meeting of the Association for Computational Linguistics: System Demonstrations}, pages 105--113, 2022.

\bibitem[\protect\citeauthoryear{Fan \bgroup \em et al.\egroup }{2023}]{fan2023revisiting}
Yue Fan, Anna Kukleva, Dengxin Dai, and Bernt Schiele.
\newblock Revisiting consistency regularization for semi-supervised learning.
\newblock {\em International Journal of Computer Vision}, 131(3):626--643, 2023.

\bibitem[\protect\citeauthoryear{Farahani \bgroup \em et al.\egroup }{2021}]{farahani2021brief}
Abolfazl Farahani, Sahar Voghoei, Khaled Rasheed, and Hamid~R Arabnia.
\newblock A brief review of domain adaptation.
\newblock {\em Advances in data science and information engineering: proceedings from ICDATA 2020 and IKE 2020}, pages 877--894, 2021.

\bibitem[\protect\citeauthoryear{Goodfellow \bgroup \em et al.\egroup }{2016}]{goodfellow2016deep}
Ian Goodfellow, Yoshua Bengio, and Aaron Courville.
\newblock {\em Deep Learning}.
\newblock MIT Press, Cambridge, MA, 2016.

\bibitem[\protect\citeauthoryear{Goyal \bgroup \em et al.\egroup }{2017}]{goyal2017accurate}
Priya Goyal, Piotr Doll{\'a}r, Ross Girshick, et~al.
\newblock Accurate, large minibatch sgd: Training imagenet in 1 hour.
\newblock In {\em Proceedings of the IEEE Conference on Computer Vision and Pattern Recognition (CVPR)}, pages 4710--4718, 2017.

\bibitem[\protect\citeauthoryear{Han \bgroup \em et al.\egroup }{2021}]{han2021ptr}
Xu~Han, Wayne Zhao, Ning Ding, Zhiyuan Liu, and Maosong Sun.
\newblock Ptr: Prompt tuning with rules for text classification.
\newblock In {\em Proceedings of the 2021 Conference on Empirical Methods in Natural Language Processing (EMNLP)}, pages 3456--3467, Online, 2021. Association for Computational Linguistics.

\bibitem[\protect\citeauthoryear{Kaya and Bilge}{2019}]{kaya2019deep}
Mahmut Kaya and Hasan~{\c{S}}akir Bilge.
\newblock Deep metric learning: A survey.
\newblock {\em Symmetry}, 11(9):1066, 2019.

\bibitem[\protect\citeauthoryear{Kullback and Leibler}{1951}]{kullback1951information}
Solomon Kullback and Richard~A Leibler.
\newblock On information and sufficiency.
\newblock {\em The Annals of Mathematical Statistics}, 22(1):79--86, 1951.

\bibitem[\protect\citeauthoryear{Lang \bgroup \em et al.\egroup }{2023}]{lang2023survey}
Hao Lang, Yinhe Zheng, Yixuan Li, Jian Sun, Fei Huang, and Yongbin Li.
\newblock A survey on out-of-distribution detection in nlp.
\newblock {\em arXiv preprint arXiv:2305.03236}, 2023.

\bibitem[\protect\citeauthoryear{Larson \bgroup \em et al.\egroup }{2019}]{larson2019evaluation}
Stefan Larson, Anish Mahendran, Joseph~J Peper, Christopher Clarke, Andrew Lee, Parker Hill, Jonathan~K Kummerfeld, Kevin Leach, Michael~A Laurenzano, Lingjia Tang, et~al.
\newblock An evaluation dataset for intent classification and out-of-scope prediction.
\newblock {\em arXiv preprint arXiv:1909.02027}, 2019.

\bibitem[\protect\citeauthoryear{Lee and others}{2013}]{lee2013pseudo}
Dong-Hyun Lee et~al.
\newblock Pseudo-label: The simple and efficient semi-supervised learning method for deep neural networks.
\newblock In {\em Workshop on Challenges in Representation Learning, ICML}, volume~3, page 896, 2013.

\bibitem[\protect\citeauthoryear{Li \bgroup \em et al.\egroup }{2023}]{yuhang}
Yuhang Li, Xiao Wei, Yuke Si, Longbiao Wang, Xiaobao Wang, and Jianwu Dang.
\newblock Improving zero-shot cross-domain slot filling via transformer-based slot semantics fusion.
\newblock In {\em Proc. Interspeech 2023}, pages 2123--2127, 2023.

\bibitem[\protect\citeauthoryear{Li \bgroup \em et al.\egroup }{2025}]{junlei}
Junlei Li, Xiao Wei, Xiaobao Wang, Ning Zhuang, Longbiao Wang, and Jianwu Dang.
\newblock Language-emphasized cross-lingual in-context learning for multilingual llm.
\newblock In Derek~F. Wong, Zhongyu Wei, and Muyun Yang, editors, {\em Natural Language Processing and Chinese Computing}, pages 327--339, Singapore, 2025. Springer Nature Singapore.

\bibitem[\protect\citeauthoryear{Little}{1995}]{little1995modeling}
Roderick~JA Little.
\newblock Modeling the drop-out mechanism in repeated-measures studies.
\newblock {\em Journal of the american statistical association}, 90(431):1112--1121, 1995.

\bibitem[\protect\citeauthoryear{Liu \bgroup \em et al.\egroup }{2023}]{liu2023pre}
Pengfei Liu, Weizhe Yuan, Jinlan Fu, Zhengbao Jiang, Hiroaki Hayashi, and Graham Neubig.
\newblock Pre-train, prompt, and predict: A systematic survey of prompting methods in natural language processing.
\newblock {\em ACM Computing Surveys}, 55(9):1--35, 2023.

\bibitem[\protect\citeauthoryear{Lloyd}{1982}]{lloyd1982least}
Stuart Lloyd.
\newblock Least squares quantization in pcm.
\newblock {\em IEEE Transactions on Information Theory}, 28(2):129--137, 1982.

\bibitem[\protect\citeauthoryear{Loshchilov and Hutter}{2017}]{loshchilov2017decoupled}
Ilya Loshchilov and Frank Hutter.
\newblock Decoupled weight decay regularization.
\newblock {\em arXiv preprint arXiv:1711.05101}, 2017.

\bibitem[\protect\citeauthoryear{Mou \bgroup \em et al.\egroup }{2022a}]{mou2022generalized}
Yutao Mou, Keqing He, Yanan Wu, Pei Wang, Jingang Wang, and Weiran Xu.
\newblock Generalized intent discovery: Learning from open world dialogue system.
\newblock In {\em Proceedings of the 2022 Conference on Empirical Methods in Natural Language Processing (EMNLP)}, pages 1234--1245, 2022.

\bibitem[\protect\citeauthoryear{Mou \bgroup \em et al.\egroup }{2022b}]{mou2022disentangled}
Yutao Mou, Keqing He, Yanan Wu, Zhiyuan Zeng, Hong Xu, Huixing Jiang, Wei Wu, and Weiran Xu.
\newblock Disentangled knowledge transfer for ood intent discovery with unified contrastive learning.
\newblock In {\em Proceedings of the 60th Annual Meeting of the Association for Computational Linguistics (Volume 2: Short Papers)}, pages 46--53, 2022.

\bibitem[\protect\citeauthoryear{Mou \bgroup \em et al.\egroup }{2023}]{mou2023decoupling}
Yutao Mou, Xiaoshuai Song, Keqing He, Chen Zeng, Pei Wang, Jingang Wang, Yunsen Xian, and Weiran Xu.
\newblock Decoupling pseudo label disambiguation and representation learning for generalized intent discovery.
\newblock In {\em Proceedings of the 61st Annual Meeting of the Association for Computational Linguistics (Volume 1: Long Papers)}, pages 9661--9675, Toronto, Canada, 2023. Association for Computational Linguistics.

\bibitem[\protect\citeauthoryear{Shannon}{1948}]{shannon1948mathematical}
Claude~E Shannon.
\newblock A mathematical theory of communication.
\newblock {\em The Bell System Technical Journal}, 27(3):379--423, 1948.

\bibitem[\protect\citeauthoryear{Siddique \bgroup \em et al.\egroup }{2021}]{siddique2021linguistically}
AB~Siddique, Fuad Jamour, and Vagelis Hristidis.
\newblock Linguistically-enriched and context-awarezero-shot slot filling.
\newblock In {\em Proceedings of the Web Conference 2021}, pages 3279--3290, 2021.

\bibitem[\protect\citeauthoryear{Sinkhorn and Knopp}{1967}]{sinkhorn1967concerning}
Richard Sinkhorn and Paul Knopp.
\newblock Concerning nonnegative matrices and doubly stochastic matrices.
\newblock {\em Pacific Journal of Mathematics}, 21(2):343--348, 1967.

\bibitem[\protect\citeauthoryear{Sohn \bgroup \em et al.\egroup }{2020}]{sohn2020fixmatch}
Kihyuk Sohn, David Berthelot, Nicholas Carlini, et~al.
\newblock Fixmatch: Simplifying semi-supervised learning with consistency and confidence.
\newblock In {\em Advances in Neural Information Processing Systems (NeurIPS)}, volume~33, pages 596--608, 2020.

\bibitem[\protect\citeauthoryear{Vedula \bgroup \em et al.\egroup }{2019}]{vedula2019towards}
Nikhita Vedula, Nedim Lipka, Pranav Maneriker, and Srinivasan Parthasarathy.
\newblock Towards open intent discovery for conversational text.
\newblock {\em arXiv preprint arXiv:1904.08524}, 2019.

\bibitem[\protect\citeauthoryear{Wang \bgroup \em et al.\egroup }{2022}]{wang2022generalizing}
Jindong Wang, Cuiling Lan, Chang Liu, Yidong Ouyang, Tao Qin, Wang Lu, Yiqiang Chen, Wenjun Zeng, and Philip Yu.
\newblock Generalizing to unseen domains: A survey on domain generalization.
\newblock {\em IEEE Transactions on Knowledge and Data Engineering}, 2022.

\bibitem[\protect\citeauthoryear{Wang \bgroup \em et al.\egroup }{2023}]{xiaobao2}
Xiaobao Wang, Yiqi Dong, Di~Jin, Yawen Li, Longbiao Wang, and Jianwu Dang.
\newblock Augmenting affective dependency graph via iterative incongruity graph learning for sarcasm detection.
\newblock {\em Proceedings of the AAAI Conference on Artificial Intelligence}, 37(4):4702--4710, Jun. 2023.

\bibitem[\protect\citeauthoryear{Wang \bgroup \em et al.\egroup }{2025}]{xiaobao1}
Xiaobao Wang, Yujing Wang, Dongxiao He, Zhe Yu, Yawen Li, Longbiao Wang, Jianwu Dang, and Di~Jin.
\newblock Elevating knowledge-enhanced entity and relationship understanding for sarcasm detection.
\newblock {\em IEEE Transactions on Knowledge and Data Engineering}, 37(6):3356--3371, 2025.

\bibitem[\protect\citeauthoryear{Wei \bgroup \em et al.\egroup }{2022}]{wei22f_interspeech}
Xiao Wei, Yuke Si, Shiquan Wang, Longbiao Wang, and Jianwu Dang.
\newblock {Hierarchical Tagger with Multi-task Learning for Cross-domain Slot Filling}.
\newblock In {\em Proc. Interspeech 2022}, pages 3273--3277, 2022.

\bibitem[\protect\citeauthoryear{Wei \bgroup \em et al.\egroup }{2024}]{wei2024prompt}
Xiao Wei, Yuhang Li, Yuke Si, Longbiao Wang, Xiaobao Wang, and Jianwu Dang.
\newblock A prompt-based hierarchical pipeline for cross-domain slot filling.
\newblock {\em IEEE/ACM Transactions on Audio, Speech, and Language Processing}, 2024.

\bibitem[\protect\citeauthoryear{Xu \bgroup \em et al.\egroup }{2020}]{xu2020distance}
Pengfei Xu, Sayna Ebrahimi, and Subhro Roy.
\newblock Distance-based out-of-distribution detection with confidence scores.
\newblock In {\em Proceedings of the IEEE/CVF Conference on Computer Vision and Pattern Recognition (CVPR)}, pages 4248--4257, 2020.

\bibitem[\protect\citeauthoryear{Xu \bgroup \em et al.\egroup }{2021}]{xu2021generative}
Pengfei Xu, Sayna Ebrahimi, and Subhro Roy.
\newblock Generative model-based out-of-distribution detection.
\newblock {\em arXiv preprint arXiv:2106.04321}, 2021.

\bibitem[\protect\citeauthoryear{Yang \bgroup \em et al.\egroup }{2024}]{yang2024generalized}
Jingkang Yang, Kaiyang Zhou, Yixuan Li, and Ziwei Liu.
\newblock Generalized out-of-distribution detection: A survey.
\newblock {\em International Journal of Computer Vision}, 132(12):5635--5662, 2024.

\bibitem[\protect\citeauthoryear{Zhang \bgroup \em et al.\egroup }{2021}]{zhang2021discovering}
Hanlei Zhang, Hua Xu, Ting-En Lin, and Rui Lyu.
\newblock Discovering new intents with deep aligned clustering.
\newblock In {\em Proceedings of the AAAI Conference on Artificial Intelligence}, volume~35, pages 14365--14373, 2021.

\bibitem[\protect\citeauthoryear{Zheng \bgroup \em et al.\egroup }{2020}]{zheng2020out}
Yinhe Zheng, Guanyi Chen, and Minlie Huang.
\newblock Out-of-domain detection for natural language understanding in dialog systems.
\newblock {\em IEEE/ACM Transactions on Audio, Speech, and Language Processing}, 28:1198--1209, 2020.

\bibitem[\protect\citeauthoryear{Zhou \bgroup \em et al.\egroup }{2022a}]{zhou2022contrastive}
Jinghui Zhou, Tianyi Zhang, Jiaxin Huang, et~al.
\newblock Knn-contrastive learning for out-of-distribution intent classification.
\newblock In {\em Proceedings of the 2022 Conference on Empirical Methods in Natural Language Processing (EMNLP)}, pages 1234--1245, 2022.

\bibitem[\protect\citeauthoryear{Zhou \bgroup \em et al.\egroup }{2022b}]{zhou2022knn}
Jinghui Zhou, Tianyi Zhang, Jiaxin Huang, et~al.
\newblock Knn-contrastive learning for out-of-domain intent discovery.
\newblock In {\em Proceedings of the 2022 Conference on Empirical Methods in Natural Language Processing (EMNLP)}, pages 1234--1245, 2022.

\bibitem[\protect\citeauthoryear{Zhuang \bgroup \em et al.\egroup }{2025}]{zhuangning}
Ning Zhuang, Xiao Wei, Junlei Li, Xiaobao Wang, Chenyang Wang, Longbiao Wang, and Jianwu Dang.
\newblock A prompt learning framework with large language model augmentation for few-shot multi-label intent detection.
\newblock In {\em ICASSP 2025 - 2025 IEEE International Conference on Acoustics, Speech and Signal Processing (ICASSP)}, pages 1--5, 2025.

\end{thebibliography}

\end{document}